\documentclass[10pt, conference, letterpaper]{IEEEtran}
\IEEEoverridecommandlockouts
\usepackage{cite}
\usepackage{amsmath,amssymb,amsfonts}
\usepackage{graphicx}
\usepackage{textcomp}
\usepackage{xcolor}
\usepackage{algpseudocode}
\usepackage{algorithm}
\usepackage{subcaption}
\def\BibTeX{{\rm B\kern-.05em{\sc i\kern-.025em b}\kern-.08em
    T\kern-.1667em\lower.7ex\hbox{E}\kern-.125emX}}

\newcommand{\fedavg}{\textsf{FedAvg}}
\newcommand{\s}{\mathbb{S}}

\newcommand{\randk}{\textsf{Random\_K}}
\newcommand{\randn}{\textsf{Random\_N}}
\newcommand{\distancen}{\textsf{Least\_distance\_N}}
\newcommand{\distancek}{\textsf{Least\_distance\_K}}
\newcommand{\stressn}{\textsf{Least\_stress\_N}}
\newcommand{\stressk}{\textsf{Least\_stress\_K}}
\newcommand{\optimaln}{\textsf{Optimal\_N}}
\newcommand{\optimalk}{\textsf{Optimal\_K}}
\newcommand{\pown}{\textsf{POW\_N}}
\newcommand{\powk}{\textsf{POW\_K}}
\newcommand{\stress}{\mathsf{Stress}}
\begin{document}

\title{On-the-fly Resource-Aware Model Aggregation for Federated Learning in Heterogeneous Edge}

\author{\IEEEauthorblockN{Hung T. Nguyen\IEEEauthorrefmark{1},
Roberto Morabito\IEEEauthorrefmark{1}, Kwang Taik Kim\IEEEauthorrefmark{2}, and Mung Chiang\IEEEauthorrefmark{2}}
\IEEEauthorblockA{\IEEEauthorrefmark{1}Princeton University, \IEEEauthorrefmark{2}Purdue University\\
Email: \IEEEauthorrefmark{1}\{hn4, roberto.morabito\}@princeton.edu, \IEEEauthorrefmark{2}\{kimkt, chiang\}@purdue.edu}}

\maketitle

\begin{abstract}
Edge computing has revolutionized the world of mobile and wireless networks world thanks to its flexible, secure, and performing characteristics. Lately, we have witnessed the increasing use of it to make more performing the deployment of machine learning (ML) techniques such as federated learning (FL). FL was debuted to improve communication efficiency compared to conventional distributed machine learning (ML). 
The original FL assumes a central aggregation server to aggregate locally optimized parameters and might bring reliability and latency issues.
In this paper, we conduct an in-depth study of strategies to replace this central server by a \emph{flying master} that is dynamically selected based on the current participants and/or available resources at every FL round of optimization.
Specifically, we compare different metrics to select this flying master and assess consensus algorithms to perform the selection.
Our results demonstrate a significant reduction of runtime using our flying master FL framework compared to the original FL from measurements results conducted in our EdgeAI testbed and over real 5G networks using an operational edge testbed.

\end{abstract}


\section{Introduction}
In the last few years, running network services or applications at the network edge became more demanded as it can provide easier access and help overcome the latency, reliability, bandwidth, or privacy issues linked to a cloud-based system. 
Edge computing also became a key enabler for the deployment of performing ML services. There is an increasing demand for running ML inference tasks at the edge, as this allows to cope with the strict latency requirements that emerging ML applications require in the context of massive and critical machine-type communications (MTC) and ultra-reliable low latency communications (URLLC). 
The more considerable interest towards edge-based ML services has also been boosted by the increasing availability of dedicated ML hardware and software tools that have been specifically designed for being suitable with the often-smaller computational capabilities of the edge. The more extensive ML tools availability has also led to increasing interest in the possibility of executing ML training tasks at the edge.
Empowering the edge with ML training capabilities can tailor those training techniques initially designed to rely on cloud services.

Federated learning (FL) \cite{mcmahan2017communication,li2020federated,wang2019adaptive,mcmahan2021advances,niknam2020federated,nguyen2020fast} (and references therein) is recently introduced as a prime ML framework for learning on edge, on local devices without sharing private data, while still benefiting from the massive and diverse available data from the federation of devices as a typical decentralized ML.
The principal design of FL includes a network of devices capable of training an ML model on local data and a central aggregation server, which routinely aggregates local models from a small set of devices to form a global model. This centralized aggregation server poses some significant issues as it is the central point of failure: \emph{(i)} if the server is unavailable due to a networking problem, e.g., under cyber-attack and/or power failure, the learning process stops and even needs to restart all over; and \emph{(ii)} the communication cost may be extremely high in practical applications on the Internet.

In this paper, we introduce a more communication-efficient FL framework to address the cumbersome centralized server setting. Earlier efforts in this direction, e.g., \cite{kim2019blockchained,qu2020decentralized,pokhrel2020federated,li2020blockchain}, employ blockchain to remove the server component. However, the employment of blockchain introduces extensive communication costs and delay due to its characteristics, rendering this approach likely unsuitable for many applications. Differently, our main idea is to dynamically select the best master node \emph{on the fly} to perform model aggregation either among the participants or the entire network at every round of FL optimization. This flying master\footnote{In our context, the term `flying master' refers to the dynamic aggregation node and not unmanned aerial vehicle (UAV) in the literature.} node selection is optimized based on communication cost and available computing resources with respect to the particular set of participants in that round. Thus, it will provide a much lower delay for both communication and aggregation.

We perform extensive experiments (Section \ref{sec:measurement}) on our EdgeAI testbed, which consists of heterogeneous edge devices 
to quantify the computational resources needed for running FL tasks, and real-world deployment of 5G testbeds \cite{5gzone} to measure communication latency over a fully operating end-to-end commercial 5G network. To pick the master node on the fly (Section \ref{subsec:general_framework}), we investigate consensus algorithms using a specified metric, termed \emph{Stress}, measuring the suitability of each candidate node (Section \ref{subsec:selection_alg}). We propose to use two algorithms based on Gossip protocol and proof-of-work consensus and perform experiments to demonstrate the better efficiency of the approaches (Section \ref{subsec:selection_alg}).
Measuring the runtime it takes to perform a number of FL rounds when each of the master node selection algorithms is used in our framework (Section \ref{sec:eval}) offers several key findings:
\emph{(i)} because fixed servers do not consider the locations of the participating devices, the central aggregation server model requires much more time than other master node selection algorithms;
\emph{(ii)} master selection algorithms based on least stress with gossip protocols and proof-of-work require mostly the same amount of runtime as the optimal selections for the same number of FL rounds because the gossip algorithm and proof-of-work puzzles are very fast compared to the local optimization time; and
\emph{(iii)} for a larger number of selected devices participating in an FL round, there is no visible difference in time consumption between the group that chooses the master among participants and the group that chooses from all devices in the network.








\section{Preliminaries}
\label{sec:prelim}
We describe the canonical FL setting with a fixed server for aggregations of local models.
In the FL setting, we have a network of $N$ devices indexed from 1 to $N$. Device $i$ possesses a local dataset $\mathcal{D}_i$ that can be used to optimize a machine learning model locally. We wish to learn a model $M$ with parameter set $\mathbf{w}$ using the available data in all the devices in a distributed manner without any device sharing its data. Hence, the optimization problem is defined as follows: $\min_{\mathbf{w}} f(\mathbf{w}) = \min_{\mathbf{w}} \frac{1}{N} \sum_{k = 1}^{N} f_k(\mathbf{w})$,
where $f(\mathbf{w})$ and $f_i\mathbf{w})$ are global and local cost functions. The local cost $f_k(\mathbf{w})$ depends on its local dataset $\mathcal{D}_k$, e.g.,
$f_k(\mathbf{w}) = \sum_{(x,y) \in \mathcal{D}_k} f_k(\mathbf{w}, (x,y))$,
where $(x,y)$ is a local data point.

In the original FL setting, there is a central server $\s{}$ in the network to perform aggregation of local models, i.e., aggregating locally updated parameters from a set of devices. A typical FL algorithm unrolls in $T$ rounds of communication and computation as described in Algorithm~\ref{alg:typical_fl}.
\begin{algorithm}
    \footnotesize
	\caption{FL algorithm with a central aggregation server}
	\label{alg:typical_fl}
	\hspace*{\algorithmicindent} \textbf{Input:} Number of participants $K$, number of rounds $T$\\
	\hspace*{\algorithmicindent} \textbf{Output:} Optimized model parameters $\mathbf{w}^T$
	\begin{algorithmic}[1] 
		\State Initialize $\mathbf{w}^0$
		\For{$t = 0, \dots, T-1$}
			\State $\s{}$ samples a set $\mathcal{S}_t$ of $K$ devices
			\State $\s{}$ sends $\mathbf{w}^t$ to all devices $k \in \mathcal{S}_t$
			\State $\mathbf{w}^{t+1} = $ \texttt{FL\_ROUND($\s, \mathbf{w}^t, \mathcal{S}_t$)}
		\EndFor
		\Procedure{\texttt{FL\_ROUND}}{$\s$, $\mathbf{w}^t, \mathcal{S}_t$}:
				\State Each device $k \in \mathcal{S}_t$ performs local optimization from $\mathbf{w}^t$ to obtain $\mathbf{w}'_k$ using its own dataset $\mathcal{D}_k$
				\State Each device $k \in \mathcal{S}_t$ sends $\mathbf{w}^{t+1}_k$ back to $\s{}$
				\State $\s{}$ aggregates $\mathbf{w}^{t+1}_k, \forall k \in \mathcal{S}_t$ to form $\mathbf{w}^{t+1}$
				\State \Return $\mathbf{w}^{t+1}$
		\EndProcedure
	\end{algorithmic}
\end{algorithm}
In round $t$, the central server first selects a small set $\mathcal{S}_t$ of $K$ devices and distributes the current model $\mathbf{w}^t$ to those selected devices (Lines~3, 7). Then, each selected device performs local optimization, e.g., commonly Stochastic Gradient Descent (SGD), starting from the received model parameters $\mathbf{w}^t$ and produces an updated parameter set $\mathbf{w}_k^{t+1}$. These parameters $\mathbf{w}_k^{t+1}, \forall k \in \mathcal{S}_t$ are sent back to the central server to be aggregated and form a new model with parameter set $\mathbf{w}^{t+1}$.

A good example of the FL algorithm is \fedavg \cite{mcmahan2017communication}, which is the first to propose FL and creates an enormous amount of follow-up works on this problem. \fedavg{} uses uniform sampling to select a set of participants in each round and employs a small number of SGD steps for local optimization. Importantly, \fedavg{} performs exceptionally well, requiring only a small number of participants in each round, e.g., $K = 10$, in a network of thousand devices, in comparison to the case where all the data are located in a single machine. In this work, we concentrate on these settings of random device selection at each round and SGD as local optimizer. Other settings in federated learning (see \cite{mcmahan2021advances}), e.g., full participation, smart device selection in each round, or different local optimizer, are feasible and left for future investigations.

\section{Measurement Methodology}
\label{sec:measurement}
We describe our experimental setup for computation and communication latency evaluations in our EdgeAI testbed and over operational 5G cellular networks, respectively.

\subsection{EdgeAI Testbed}
\begin{figure}
\includegraphics[width=0.35\textwidth]{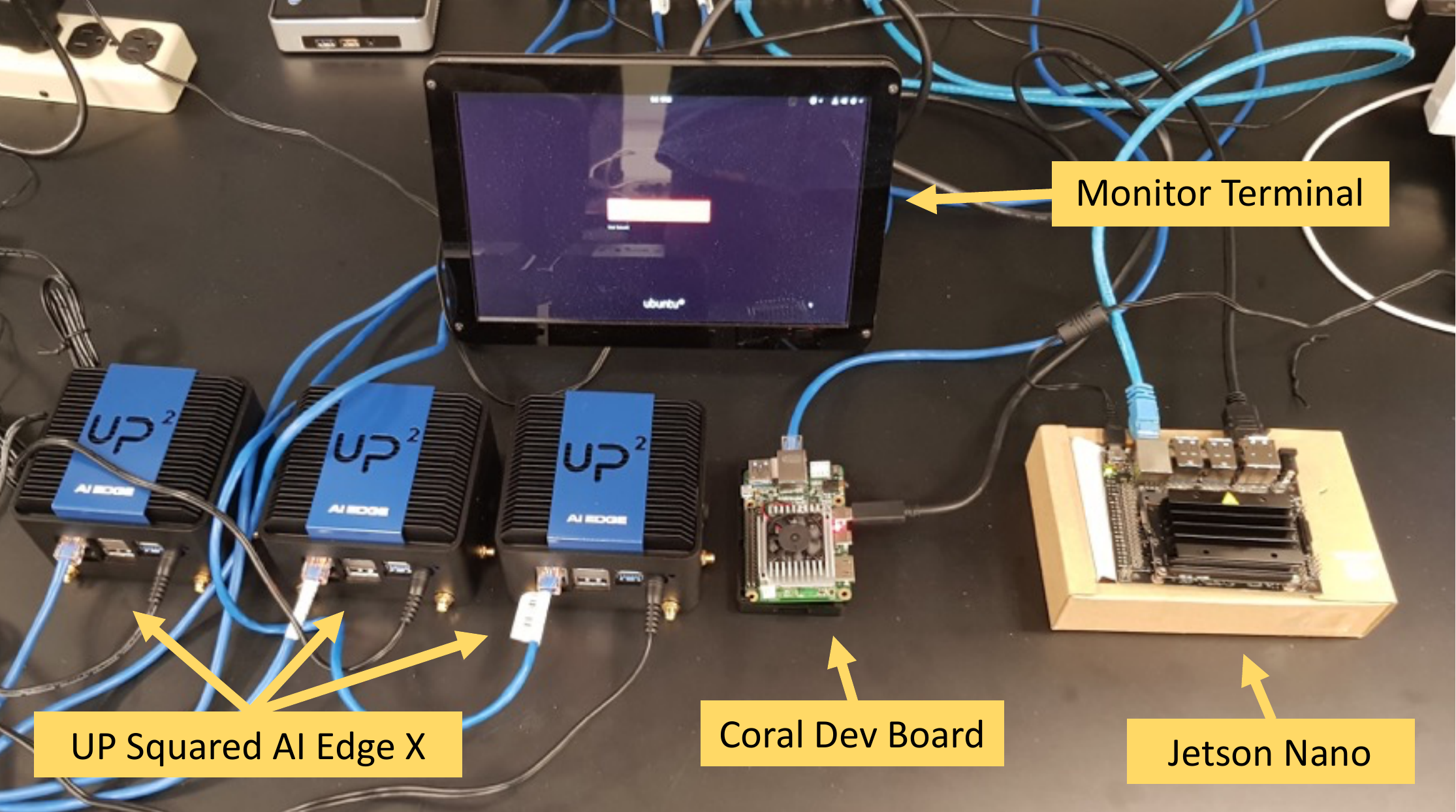}
\centering
\caption{EdgeAI testbed setup.}
\label{fig:edgeai}
\vspace{-0.15in}
\end{figure}

\begin{figure}
\includegraphics[width=0.4\textwidth]{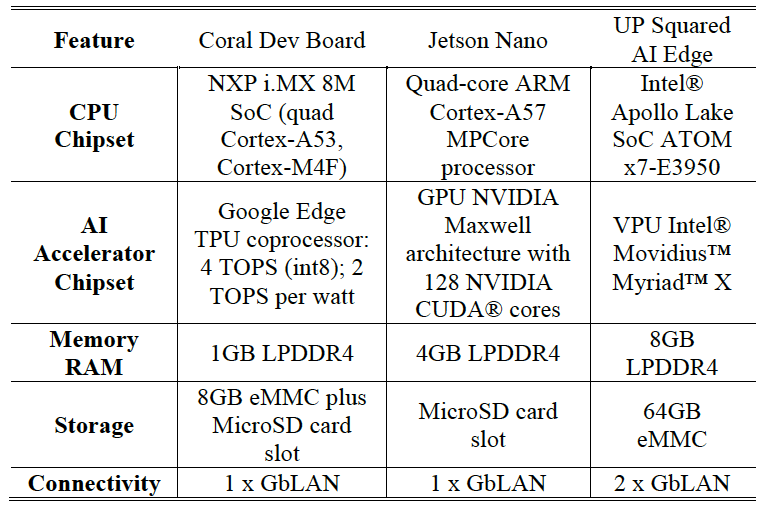}
\centering
\caption{EdgeAI testbed devices' hardware features.}
\label{fig:edgeai_hw}
\vspace{-0.22in}
\end{figure}

Our EdgeAI testbed (Fig.~\ref{fig:edgeai}) introduces edge devices' heterogeneity by using three distinct types of AI accelerators, such as Intel Movidius Myriad X VPU, Google Edge TPU, and NVIDIA 128-core Maxwell GPU. The three systems-on-chip are hosted in three different Single-Board Computers (SBCs): UP Squared AI Edge, Coral Dev Board, and NVIDIA Jetson Nano, respectively. Fig.~\ref{fig:edgeai_hw} summarizes the other hardware characteristics of the devices. It can be noticed how the SBCs have heterogeneous features also from CPU, memory RAM, and storage perspectives.

\subsection{5G Testbed}
\label{subsec:5gtestbed}
\subsubsection{Measurement Setup}
We utilize a fully operational end-to-end 5G network testbed in Indy 5G Zone \cite{5gzone} to measure the impact of various realistic radio propagation and wireline backhaul in commercial 5G networks. The testbed consists of 5G gNB operating at 39GHz sponsored by AT\&T. The architecture adopts 3GPP NSA Option 3, where UE is anchored to LTE core network via existing LTE/EPC control plane.
Our user equipment (UE) is a Samsung Galaxy S20 Ultra 5G (Qualcomm Snapdragon 865 \cite{snapdragon865}). We use a server in Utah from Cloud Lab \cite{cloudlab} and connect it through a public IP address for experiments. We measure a TCP congestion control algorithm's latency in the Linux kernel module over a 5G network by setting up a 5G phone as a TCP receiver.

\subsubsection{Data Collection and Processing}
We use iperf3 \cite{iperf3} to generate TCP traffic and RTT as a key communication latency metric. The first 10 seconds of each experiment period are excluded to reduce the starting effect when measuring RTT. We run iperf3 for 70 seconds using a congestion control algorithm, ExLL \cite{tcp-exll}. We repeated this experiment five times.

\section {Flying Master FL Framework}
\vspace{-0.03in}
In this section, we introduce our flying master FL framework. In a nutshell, instead of having a fixed centralized aggregation server, the participants in each round will pick their master node based on some criterion, which is the most suitable regarding communication cost and computation capacity. In effect, this will reduce the time it takes for every round since, given all the information of current participants, we can find a much better candidate than a predefined server in the original FL setting.

In the following subsections, we describe our general framework, discuss criteria to select the most suitable master node, and then investigate algorithms to perform the selection in different scenarios of security/privacy concerns.
\vspace{-0.05in}
\subsection{General Framework} \label{subsec:general_framework}

Our general flying master FL framework is described in detail in Algorithm~\ref{alg:flying_fl}.
\begin{algorithm}
    \footnotesize
	\caption{Flying master FL General Framework}
	\label{alg:flying_fl}
	\hspace*{\algorithmicindent} \textbf{Input:} $K$, $T$\\
	\hspace*{\algorithmicindent} \textbf{Output:} Optimized model parameters $\mathbf{w}^T$
	\begin{algorithmic}[1] 
		\State Initialize $\mathbf{w}^0$
		\State Set $\s_0$ to an  initial master node
		\For{$t = 0, \dots, T-1$}
		\State $\s_t$ samples a set $\mathcal{S}_t$ of $K$ devices
		\State $\s_t$ sends $\mathbf{w}^t$ to all devices $k \in \mathcal{S}_t$
		\State A new flying master $\s_{t+1}$ is selected based on $\mathcal{S}_t$ \Comment{by a master selection algorithm discussed in the next subsection}
		\State $\mathbf{w}^{t+1} = $ \texttt{FL\_ROUND($\s_{t+1}, \mathbf{w}^t, \mathcal{S}_t$)} \Comment{as in Alg.~\ref{alg:typical_fl}}
		\EndFor
	\end{algorithmic}
\end{algorithm}
\vspace{-0.05in}
Here a new master node is dynamically selected at every round of optimization with respect to the particular set $\mathcal{S}_t$ of participants (Line 5). This master node selection algorithm may depend on different criteria and will be discussed in the subsequent subsection. In round $t$, given a new master node $\s_{t+1}$, the following steps are similar to the FL algorithm with a central server (Line 6). The initial master node $\s_0$ can be one initiating the learning process (Line 2).

Our framework's special case is the original FL algorithm with a central aggregation server when the master node selection algorithm is to return a predefined $\s$. Thus, the flying master FL framework is a strict generalization of the original FL algorithms and provides more flexibility to improve its performance in latency or convergence time.


\subsection{Master Selection Algorithms} \label{subsec:selection_alg}
The selection of a flying master node in each round can be based on different criteria to minimize latency. Each criterion may also be implemented by different algorithms when security/privacy is of paramount concern. We investigate and compare the following criteria:
\begin{itemize}
	\item \textbf{Random selection}: This selects a node uniformly at random among the $K$ participants or all $N$ devices, denoted by \randk{} and \randn{}, respectively.
	\item \textbf{Least distance}: This criterion finds the node with the smallest sum of distances between $K$ participants and that node. Intuitively, this smallest distance node has the least communication delay to the participants. Here distances can be estimated based on coordinates of devices or communication latency. We also have two versions, \distancek{} and \distancen{}, to represent considerations among $K$ participants and all $N$ devices.
	\item \textbf{Least stress}: Taking into account how much resources are available at the nodes or their stress levels, we consider criteria based on the device's stress, namely \stressk{} and \stressn{}, with the philosophy that the less stress a device has, the more suitable the device is to serve as flying master. This stress metric reflects the amount of used resources, including CPU, memory, network bandwidth, and is calculated as follows:
	\begin{align}
		\stress = \frac{1}{\mathsf{CPU}} \times \frac{1}{\mathsf{MEM}} \times \frac{1}{\mathsf{NET}},
	\end{align}
	where $\mathsf{CPU}$ measures CPU speed (GHz), $\mathsf{MEM}$ is for available memory (GB), and $\mathsf{NET}$ assesses available network bandwidth (MB/s).
\end{itemize}
The last two criteria require computation at all the candidate nodes and possibly private information exchanges between them to pick the most suitable node to be the flying master. Thus, this selection leads to a consensus problem that involves all the candidate nodes to agree on a master with the smallest metric. We discuss two approaches to deal with our consensus problem: Gossip protocol and Proof-of-work.

\subsubsection{Gossip Protocol based Flying Master Selection} Gossip protocols can be typically used for different purposes, including cluster membership identification, cluster's node failure detection, and event broadcast \cite{8416202}, with minimal bandwidth and latency overhead \cite{rossi2014distributed}. Their characteristics and peer-to-peer design make them very suitable for distributed systems that require high scalability and reliability for system information sharing. We take advantage of this gossip-based approach for sharing the stress metric associated with each candidate for master selection during each FL round. In particular, benefiting from the event-based broadcast nature of this protocol, similarly to the Max-consensus approach in \cite{6259916}, the \textit{Flying Master Selection} is performed in the following steps:
\begin{itemize}
  \item After a gossip interval, e.g., 10ms, a node \textit{k} wakes up.
  \item Node \textit{k} broadcasts its current \textit{least stress} value to all the nodes in the candidate set ($\mathcal{S}^t$ at round $t$ or all $N$ nodes).
  \item Each node updates the least stress by taking minimum.
\end{itemize}
In the end, every node knows which node has the least stress.

\subsubsection{Proof-of-Work-based Flying Master Selection}
The issue with using gossip protocol is that the devices have to share information about their resources, which may not be desirable given growing concerns about unforeseeable security and privacy problems. We describe the second approach of using proof-of-work to perform consensus and indirectly evaluating the least distance and least stress criteria mentioned previously. It offers a nice property that devices do not share private information and fits the FL's privacy-preserving requirement. However, solving proof-of-work puzzles requires more energy from participating devices; thus, there is a trade-off between these approaches' security and energy usage.

Proof-of-work\cite{dwork1992pricing,jakobsson1999proofs,wang2003defending} involves a moderately hard but not intractable puzzle, e.g., hash inversion, usually with a parameter to control the hardness translating to the time it takes to solve the puzzle. A certain amount of computation is required to find a solution to the puzzle; however, it is effortless to verify that a solution is valid \cite{gervais2016security}. Thus, the more computational resources a device has, the faster it will be able to find a solution. In our framework, we use this property to determine which device has the most computational resources available. Thus, it is more suitable to become the flying master for the current round. Additionally, to account for the communication capacity, a device needs to send the solution to every participant to verify after solving the puzzle. Once all these are done, the node will finally become the new master node. As a result, this proof-of-work approach quantifies both computation and communication resources as a combination of least distance and least stress criteria.

\begin{algorithm}
    \footnotesize
	\caption{Proof-of-work based flying master selection}
	\label{alg:proof_of_work}
	\hspace*{\algorithmicindent} \textbf{Input:} A set of devices $\mathcal{S}$, proof-of-work puzzle $\mathfrak{p}$\\
	\hspace*{\algorithmicindent} \textbf{Output:} Flying master node $\s$
	\begin{algorithmic}[1]
		\ForAll{$k \in \mathcal{S}$ \textbf{in parallel}}
		\State $k$ runs \textsc{Solve}($\mathfrak{p}$) and \textsc{Verify}($\mathfrak{p}$) in parallel
		\EndFor
		\Procedure{Solve}{$\mathfrak{p}$}
		\State $k$ attempts to find a solution $\mathfrak{s}$ of $\mathfrak{p}$
		\State $k$ sends its' \textsf{id} and solution $\mathfrak{s}$ to other devices
		\If{$k$ receives \textsf{verified} from all other devices}
		\State $k$ declares to be the new flying master
		\EndIf
		\EndProcedure
		\Procedure{Verify}{$\mathfrak{p}$}
		\If{received \textsf{id} and solution $\mathfrak{s}$ from device $k'$}
		\State Verify $\mathfrak{s}$ along with \textsf{id} of $k'$ is valid
		\If{Verification is successful}
		\State $k$ sends message \textsf{Verified} back to $k'$
		\EndIf
		\EndIf
		\If{Observe master declaration from a verified $k'$}
		\State Halt both \textsc{Solve}($\mathfrak{p}$) and \textsc{Verify}($\mathfrak{p}$) threads
		\EndIf
		\EndProcedure
	\end{algorithmic}
\end{algorithm}
\vspace{-0.05in}
The details of our proof-of-work-based master selection algorithm are provided in Algorithm~\ref{alg:proof_of_work} and performed in every FL round. A puzzle $\mathfrak{p}$ is created uniquely for each round and each device and can be constructed from the information of current model parameters, round index, and device's id. Each device $k$ in parallel attempts to find a solution and verifies solutions from other devices (Line 2). If $k$ finds a solution $\mathfrak{s}$ before the master node has been declared, it will send the solution along with its \textsf{id} to all others to verify (Lines~5, 6). Then, if all other devices verify its solution and send back all \textsf{verified} messages, $k$ declares to be the new master node (Lines~7, 8). During solving the puzzle, each device $k$ also engages in verifying other solutions if it receives a request (Lines~12-17). All the devices will stop both finding and verifying whenever a device declares to be the new master (Lines~18-20). Note that this new master node must be already verified by every node (Line~18). Depending on the set $\mathcal{S}$ to include only current participants or all devices in the network, we term the respective selection algorithms \powk{} and \pown{} in the following evaluations.

\section{Numerical Evaluations} 
\label{sec:eval}
We evaluate our framework with different settings of master node selection criteria and algorithms, including the original central aggregation server setting, using our Edge AI and 5G testbeds described in Section~\ref{sec:measurement}.
\begin{figure}[t]
\vspace{-0.1in}
\includegraphics[width=0.45\textwidth]{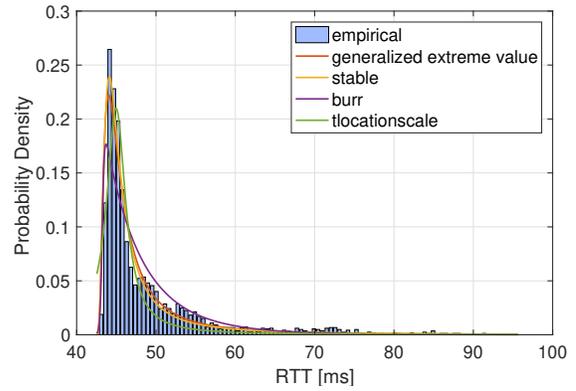}
\centering
\caption{The PDF of measured latencies over a 5G network between a 5G device and a cloud server, shown with the best-fitting distributions. In many cases we examine, the Generalized Extreme Value distribution is the best fit out of 20 common statistical distributions.}
\vspace{-0.18in}
\label{fig:PDFlatenciesGEV}
\end{figure}
\begin{figure*}[t!]
	\centering
	\includegraphics[width=0.9\linewidth]{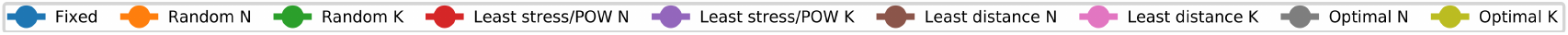}
	\vspace{-0.15in}
\end{figure*}
\begin{figure*}[t!]
	\centering
	\begin{subfigure}[b]{0.205\linewidth}
		\includegraphics[width=\linewidth]{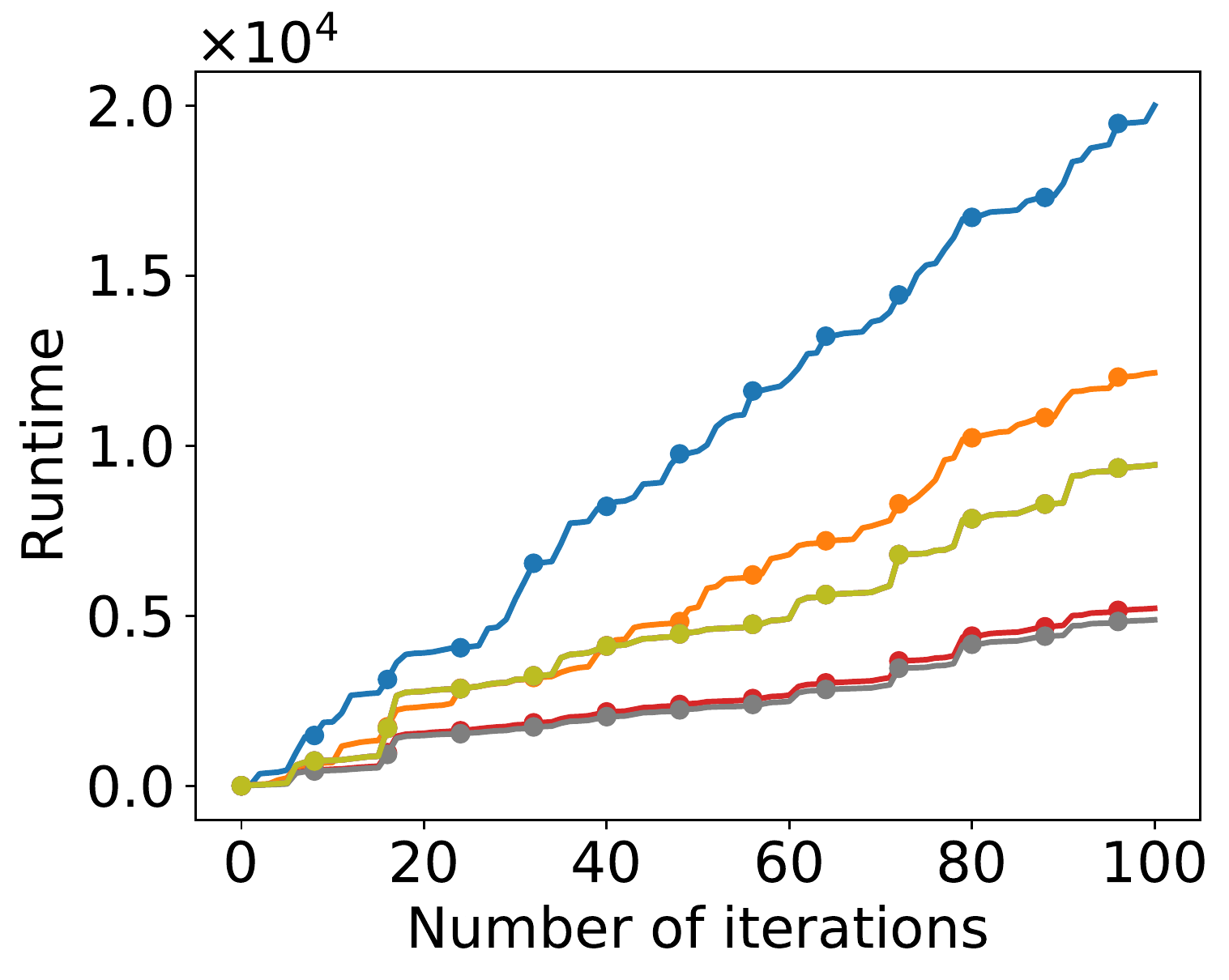}
		\caption{$K = 1$}
	\end{subfigure}
	\begin{subfigure}[b]{0.205\linewidth}
		\includegraphics[width=\linewidth]{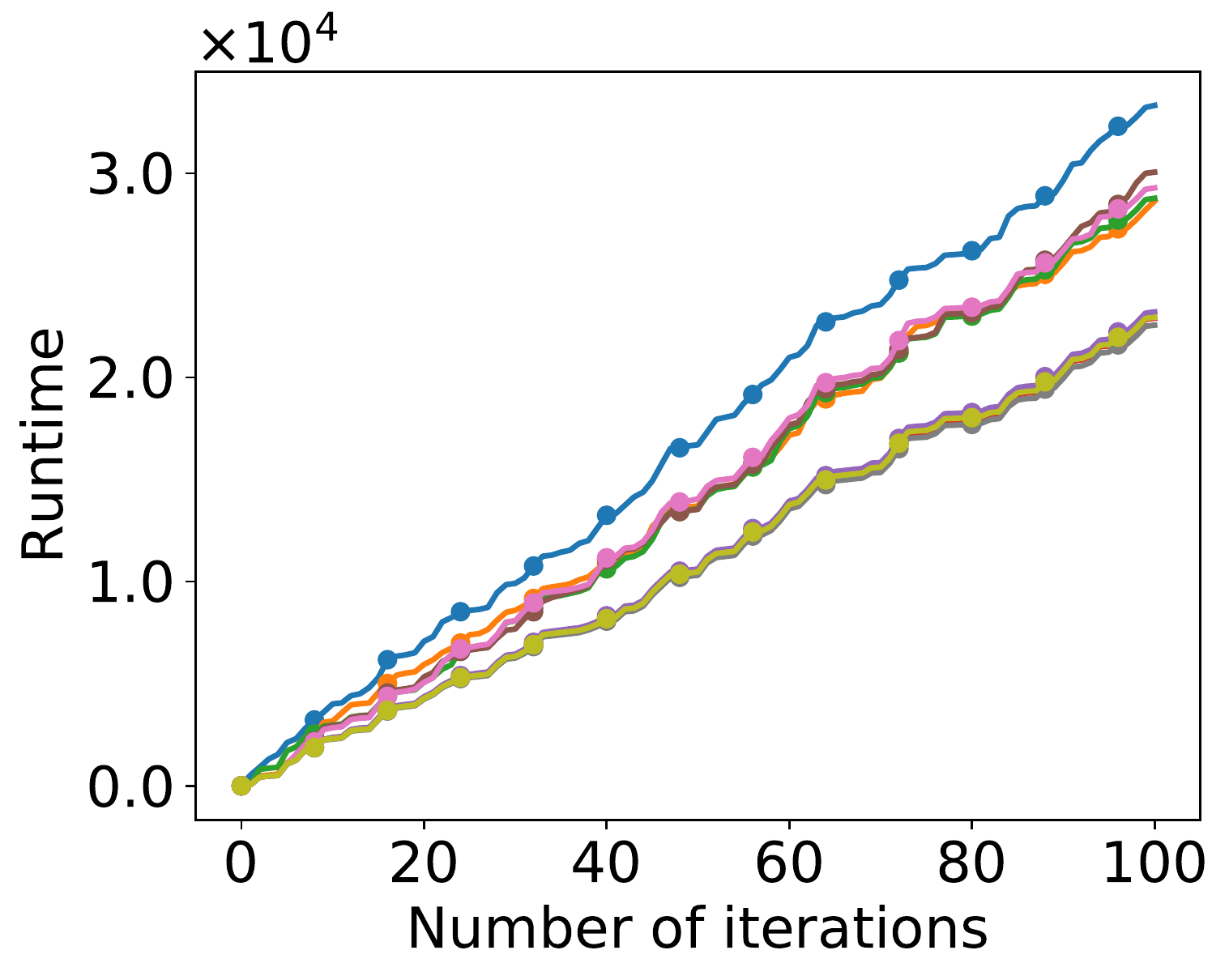}
		\caption{$K = 5$}
	\end{subfigure}
	\begin{subfigure}[b]{0.205\linewidth}
		\includegraphics[width=\linewidth]{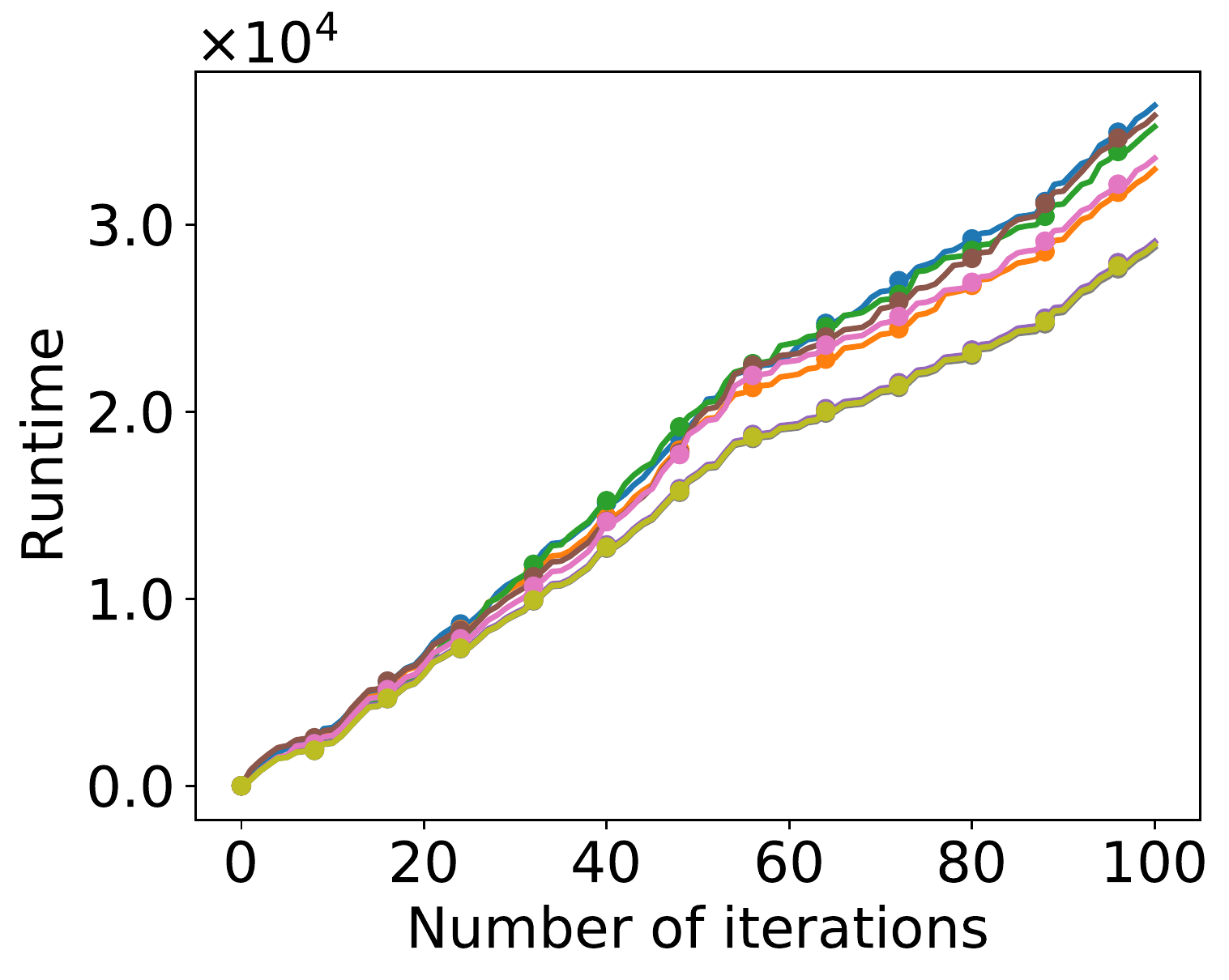}
		\caption{$K = 10$}
	\end{subfigure}
	\begin{subfigure}[b]{0.205\linewidth}
		\includegraphics[width=\linewidth]{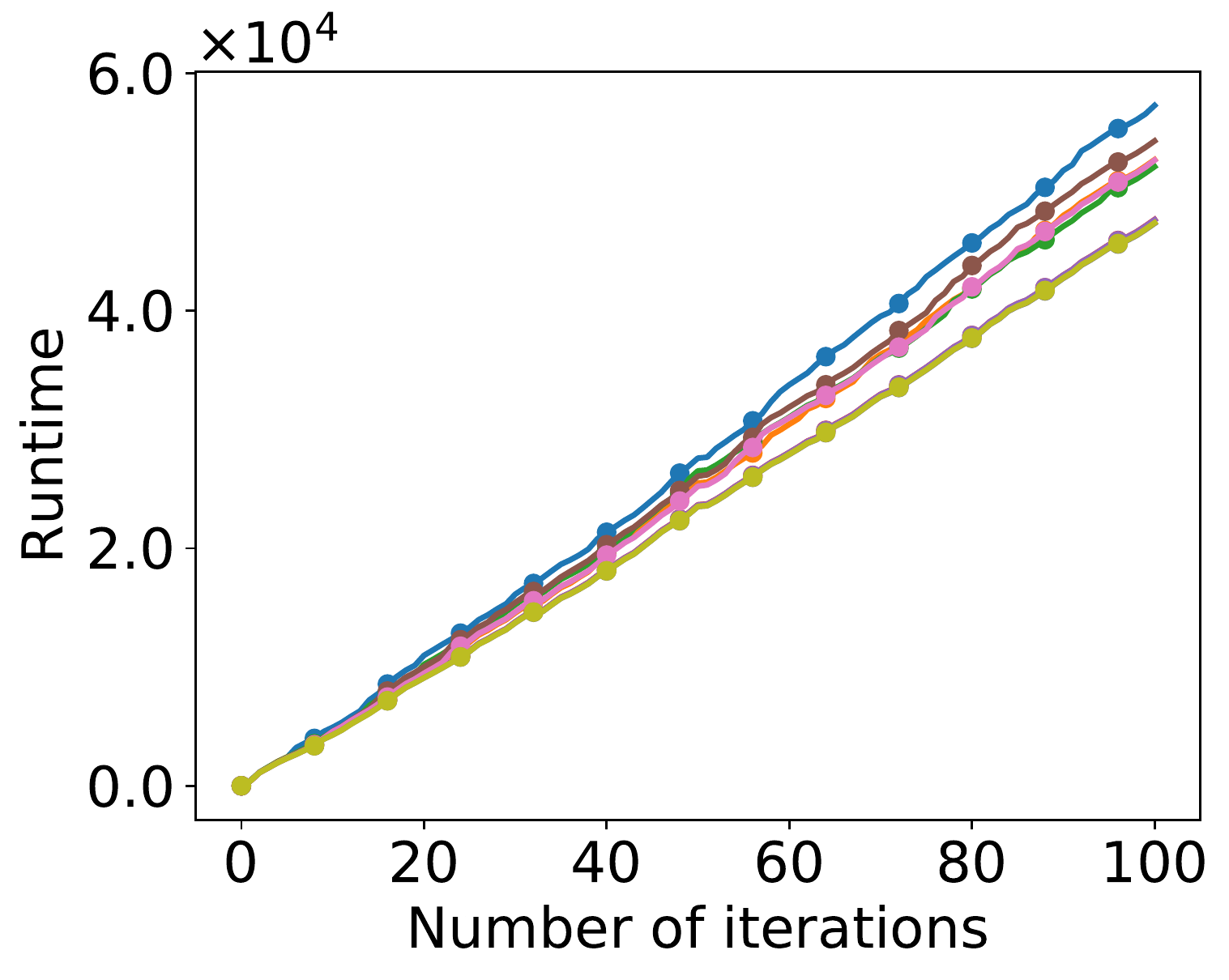}
		\caption{$K = 50$}
	\end{subfigure}
	\vspace{-0.05in}
	\caption{Wall-clock time vs. number of rounds for $N = 100$ and 2 types of devices in the network.}
	\label{fig:N100_d2}
	\vspace{-0.15in}
\end{figure*}
\begin{figure*}[!t]
	\centering
	\begin{subfigure}[b]{0.205\linewidth}
		\includegraphics[width=\linewidth]{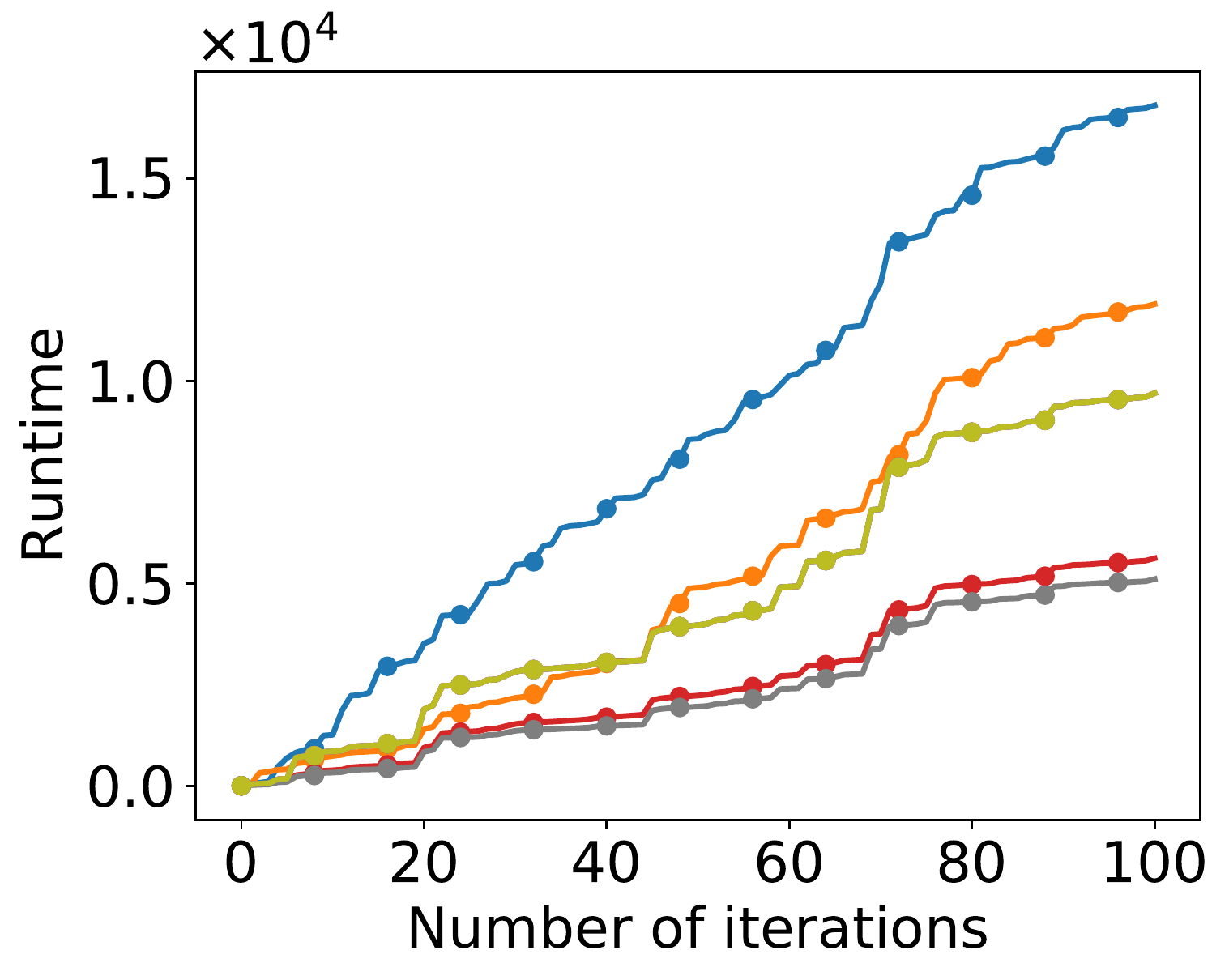}
		\caption{$K = 1$}
	\end{subfigure}
	\begin{subfigure}[b]{0.205\linewidth}
		\includegraphics[width=\linewidth]{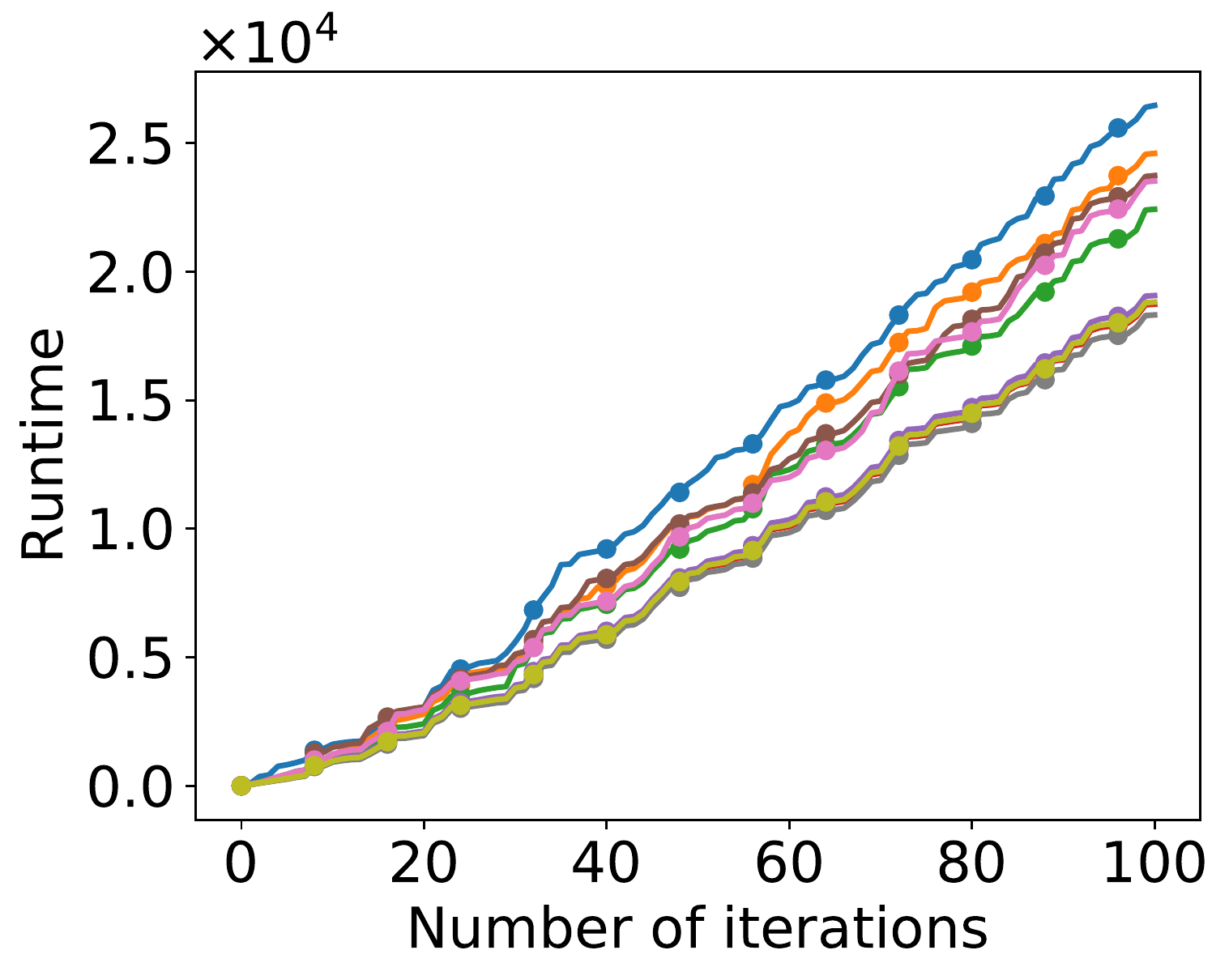}
		\caption{$K = 5$}
	\end{subfigure}
	\begin{subfigure}[b]{0.205\linewidth}
		\includegraphics[width=\linewidth]{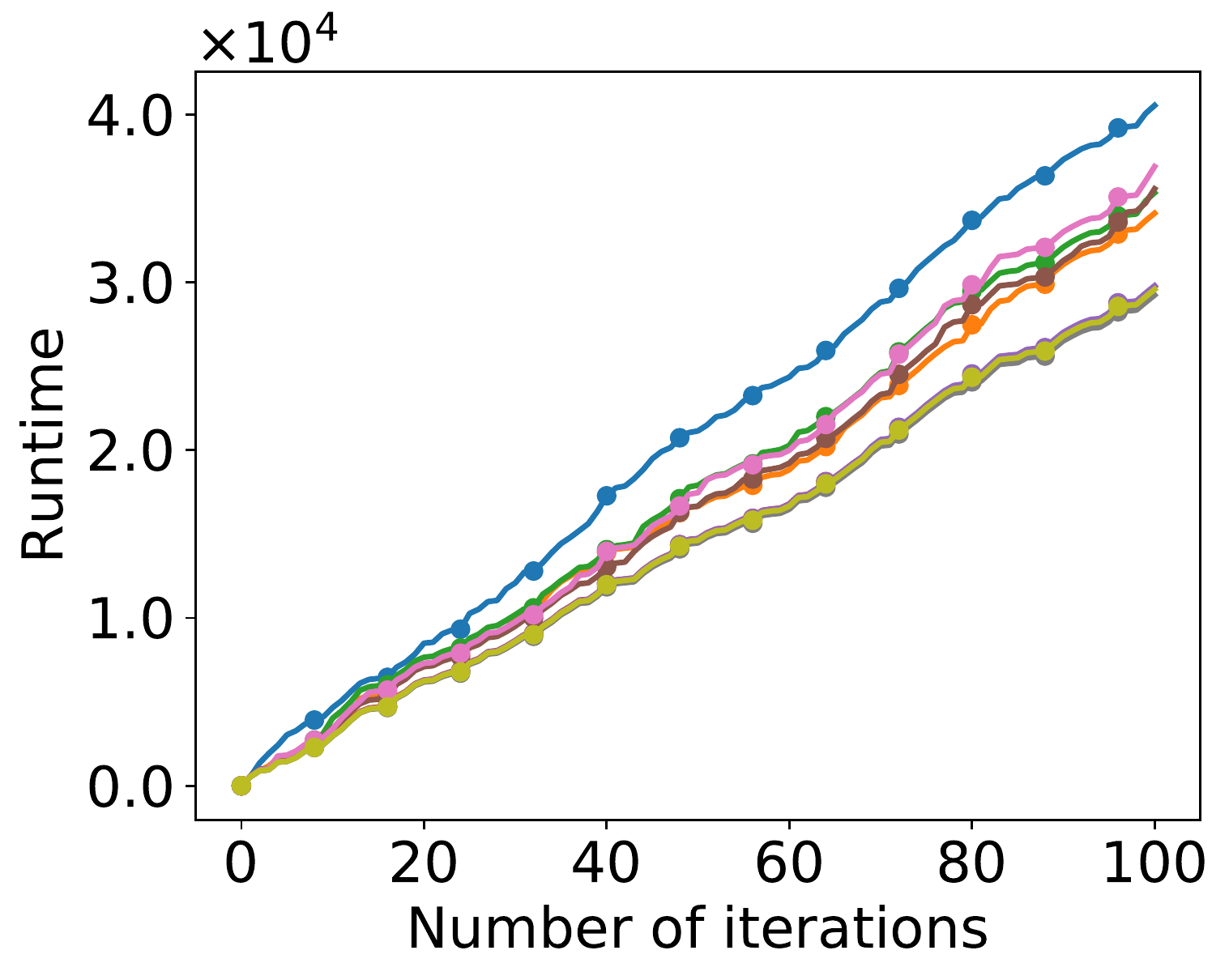}
		\caption{$K = 10$}
	\end{subfigure}
	\begin{subfigure}[b]{0.205\linewidth}
		\includegraphics[width=\linewidth]{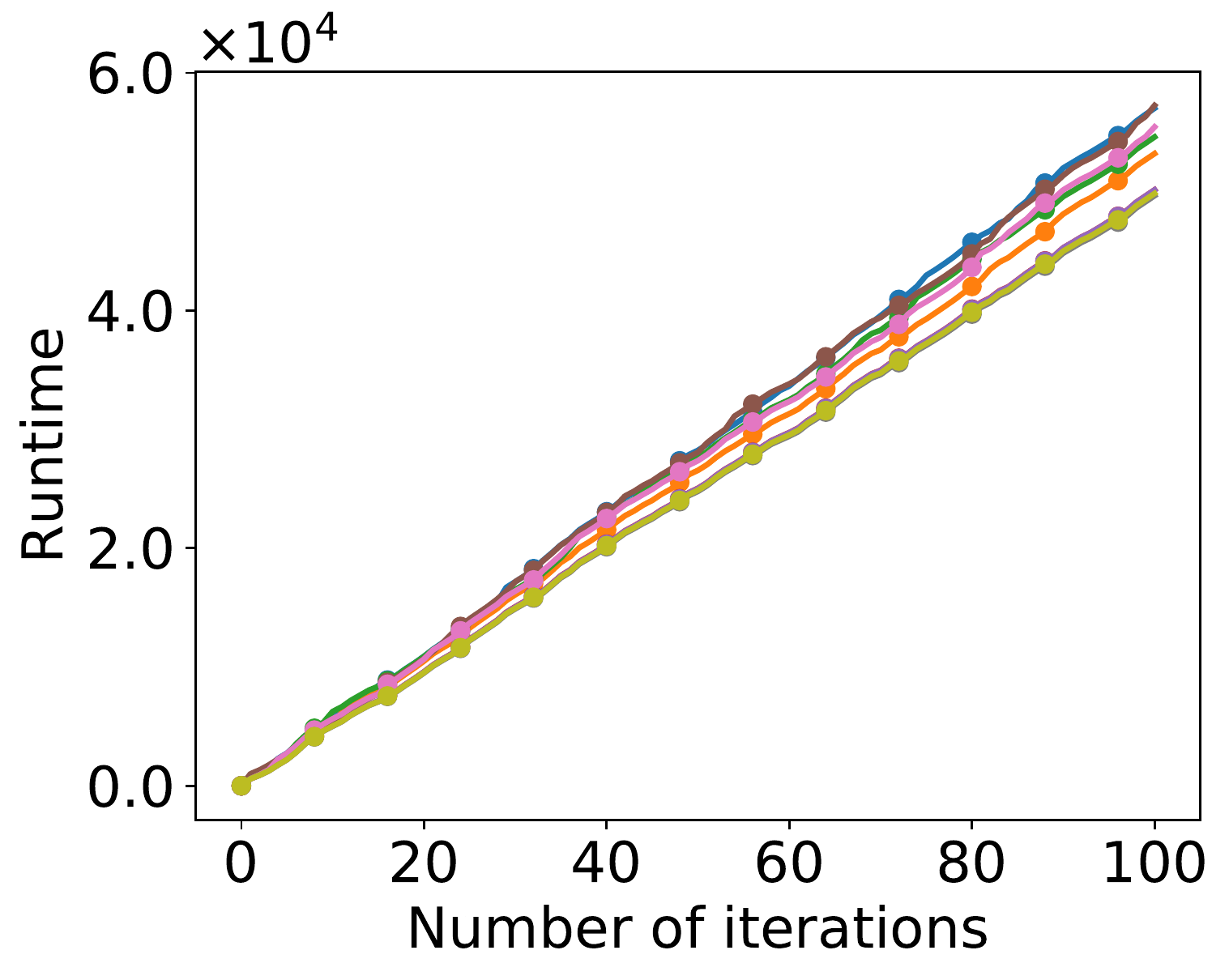}
		\caption{$K = 50$}
	\end{subfigure}
	\vspace{-0.05in}
	\caption{Wall-clock time vs. number of rounds for $N = 1000$ and 2 types of devices in the network.}
	\label{fig:N1000_d2}
	\vspace{-0.15in}
\end{figure*}
\begin{figure*}[t!]
	\centering
	\begin{subfigure}[b]{0.205\linewidth}
		\includegraphics[width=\linewidth]{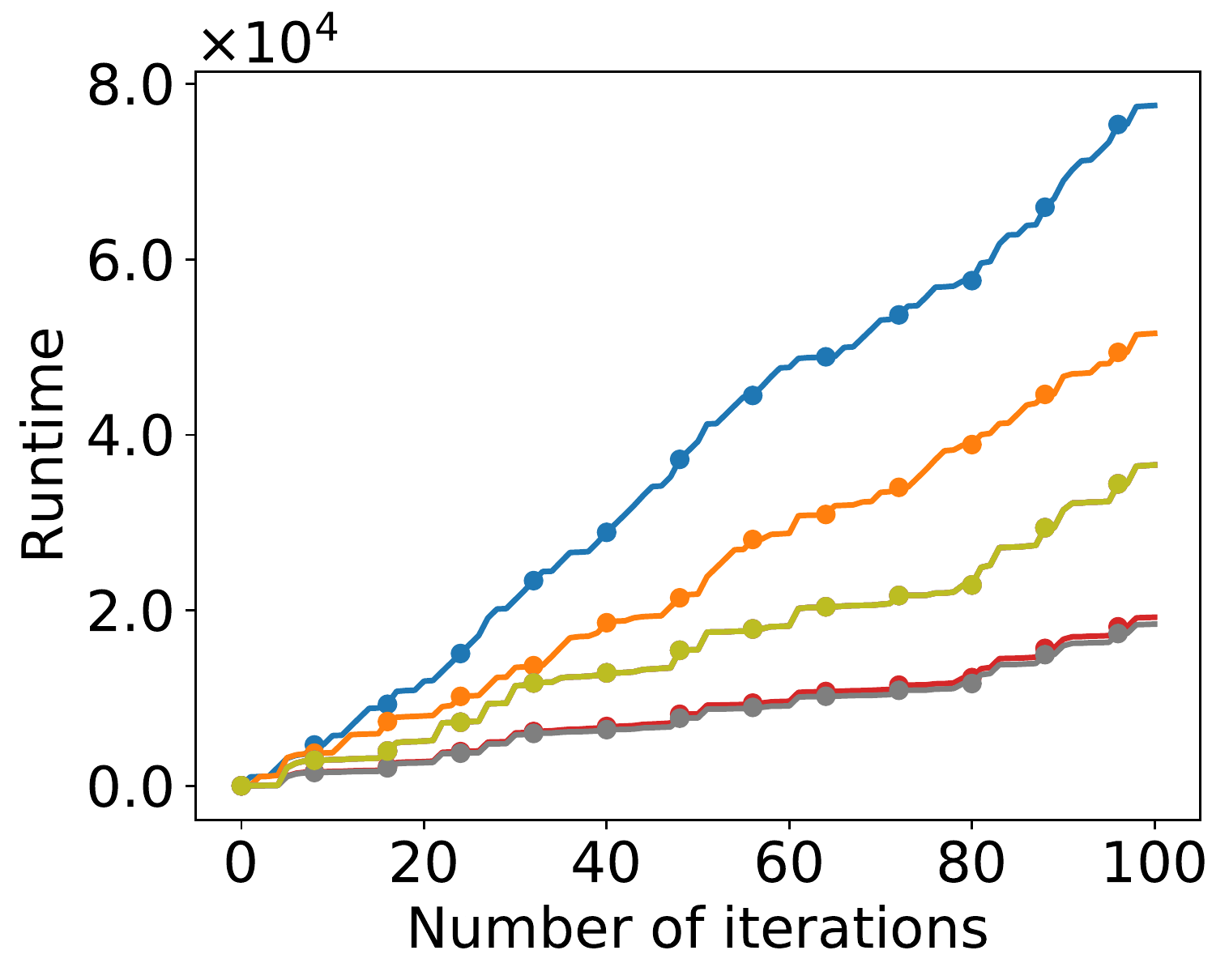}
		\caption{$K = 1$}
	\end{subfigure}
	\begin{subfigure}[b]{0.205\linewidth}
		\includegraphics[width=\linewidth]{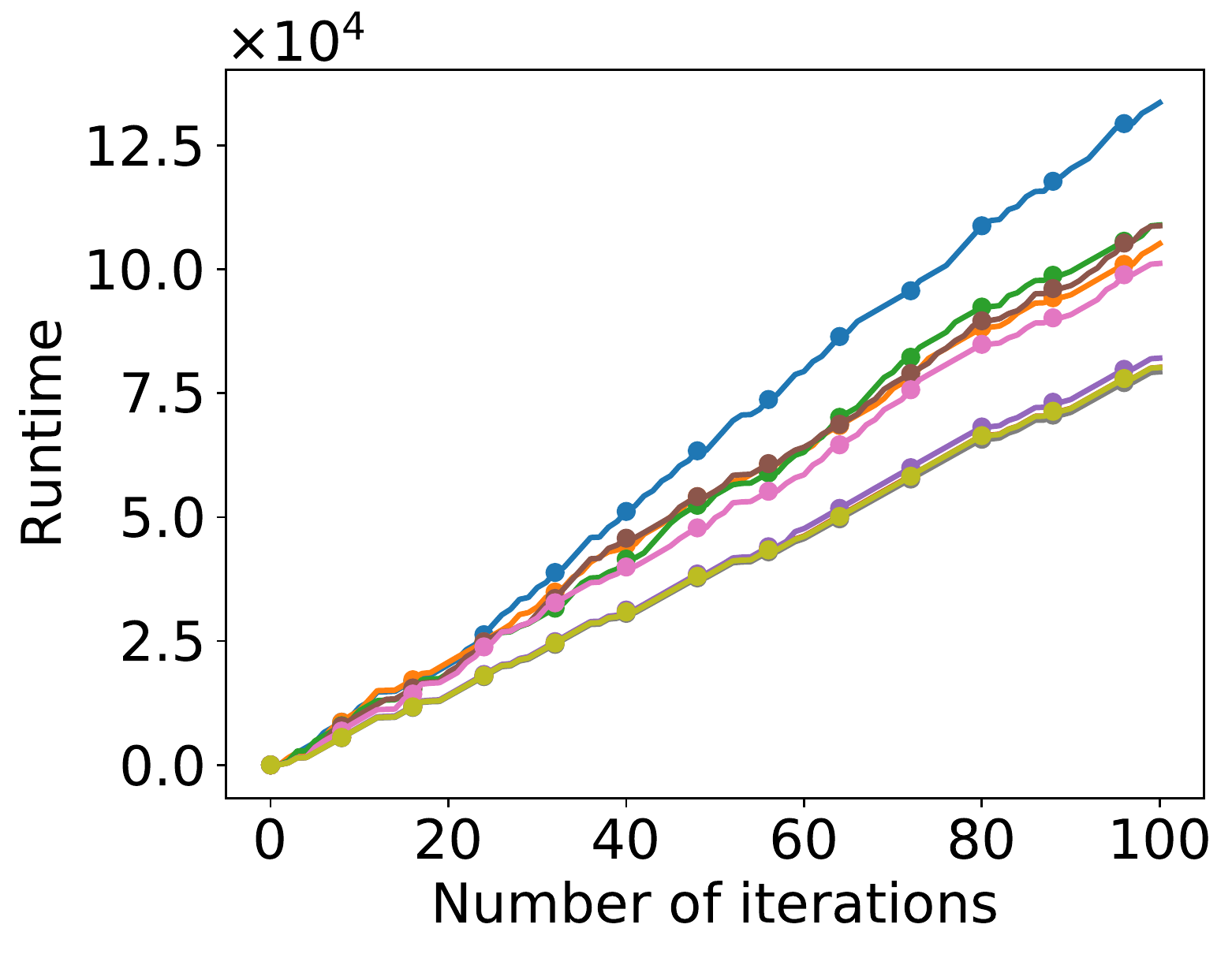}
		\caption{$K = 5$}
	\end{subfigure}
	\begin{subfigure}[b]{0.205\linewidth}
		\includegraphics[width=\linewidth]{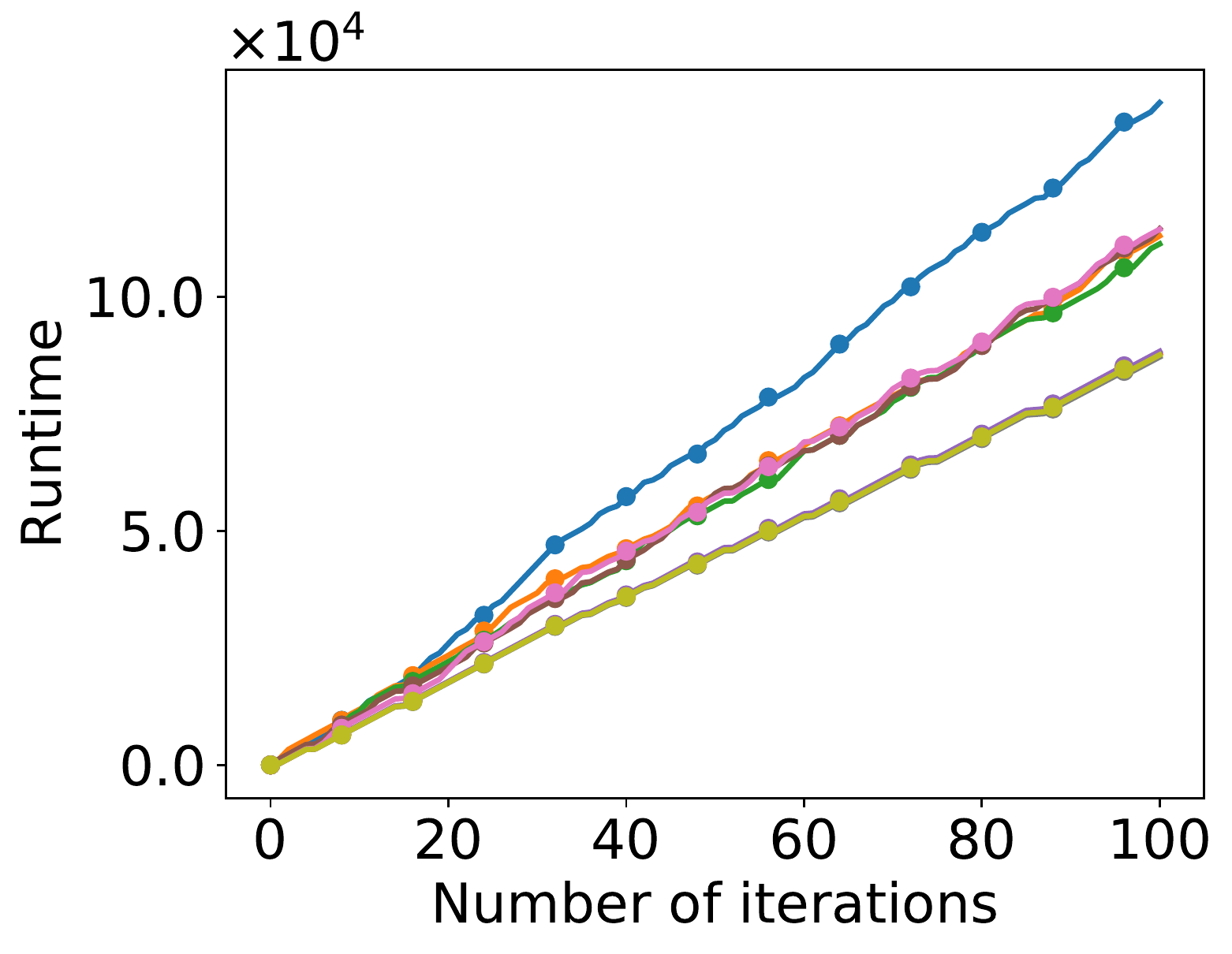}
		\caption{$K = 10$}
	\end{subfigure}
	\begin{subfigure}[b]{0.205\linewidth}
		\includegraphics[width=\linewidth]{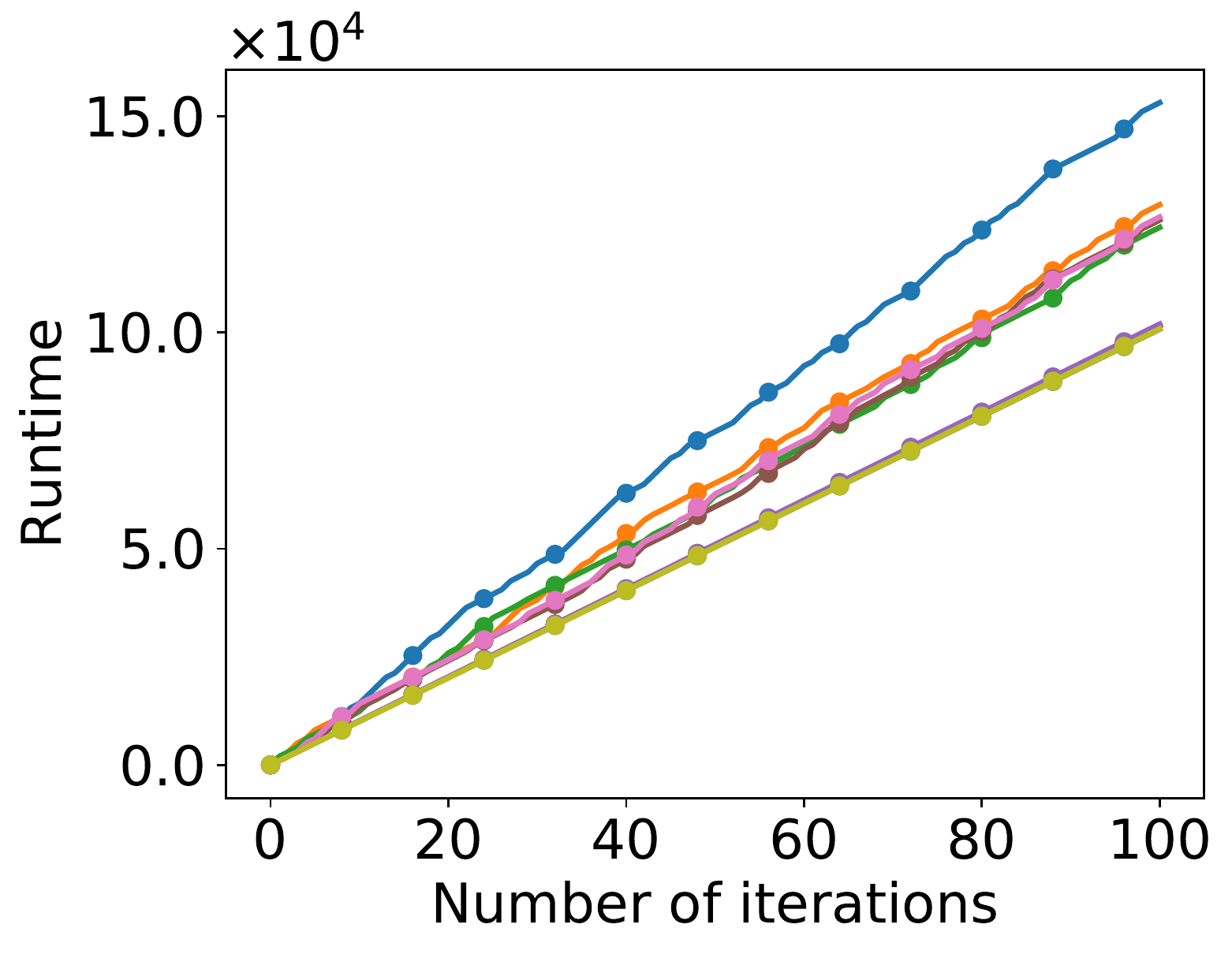}
		\caption{$K = 50$}
	\end{subfigure}
	\vspace{-0.05in}
	\caption{Wall-clock time vs. number of rounds for $N = 100$ and 4 types of devices in the network.}
	\label{fig:N100_d4}
	\vspace{-0.15in}
\end{figure*}
\begin{figure*}[!t]
	\centering
	\begin{subfigure}[b]{0.205\linewidth}
		\includegraphics[width=\linewidth]{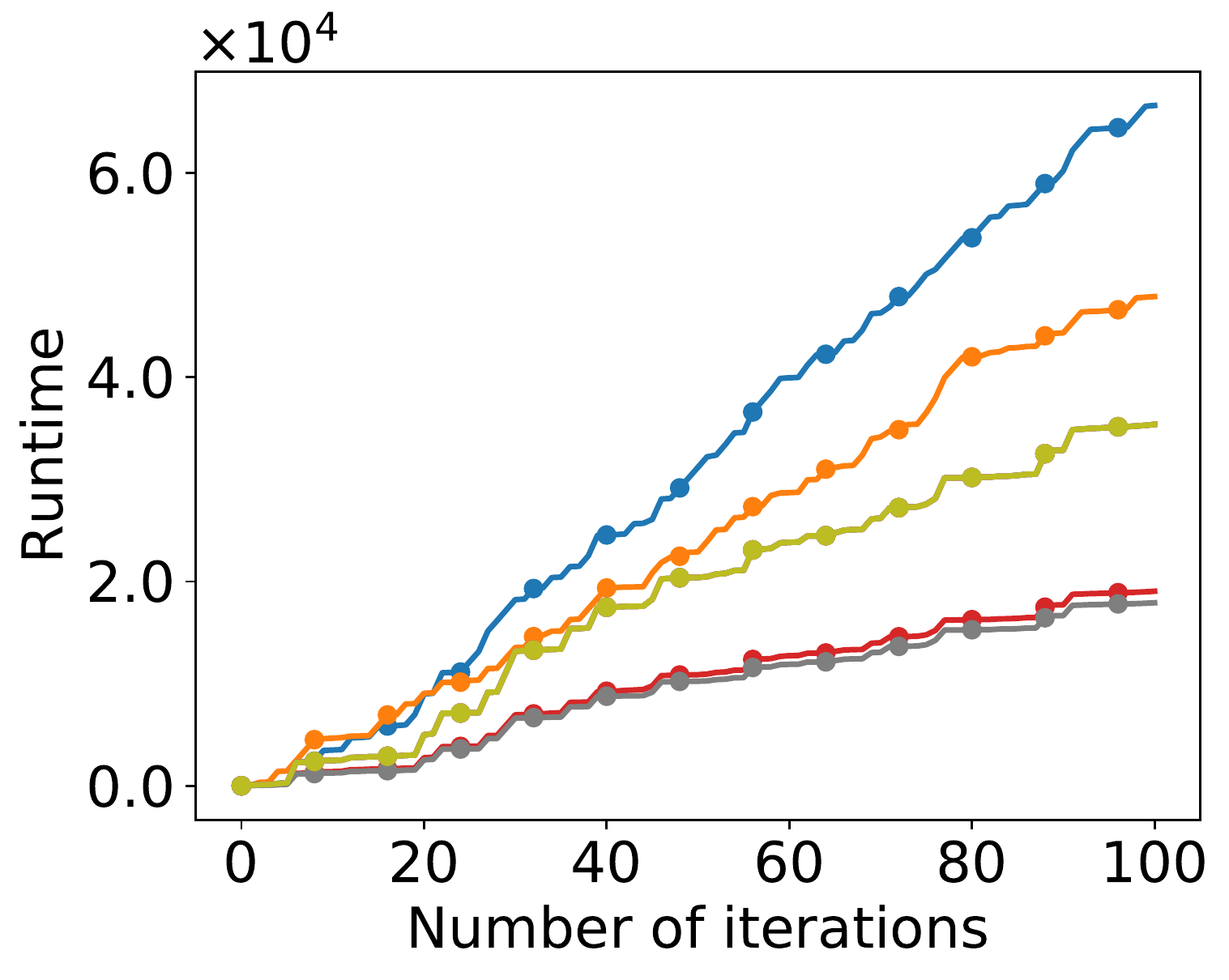}
		\caption{$K = 1$}
	\end{subfigure}
	\begin{subfigure}[b]{0.205\linewidth}
		\includegraphics[width=\linewidth]{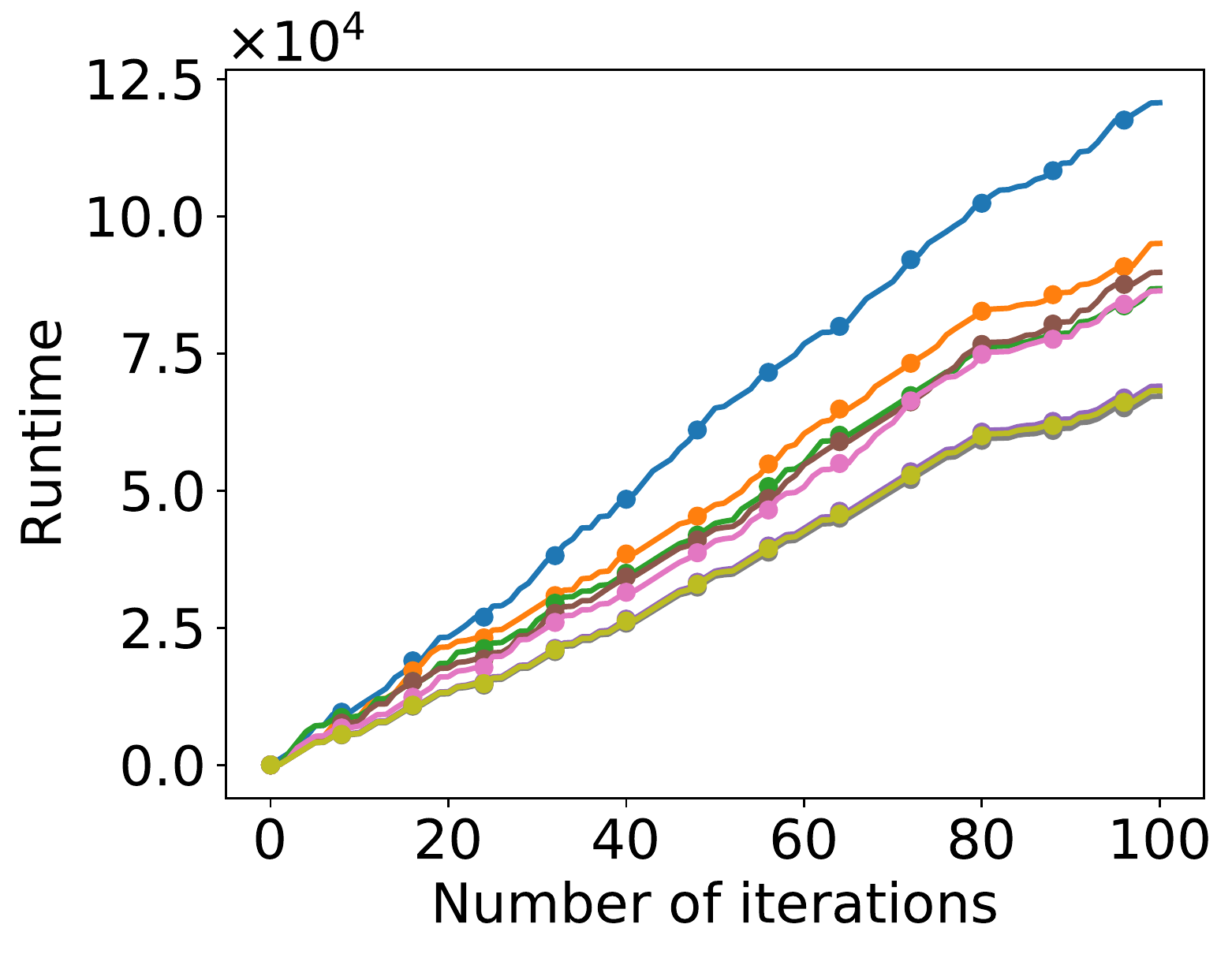}
		\caption{$K = 5$}
	\end{subfigure}
	\begin{subfigure}[b]{0.205\linewidth}
		\includegraphics[width=\linewidth]{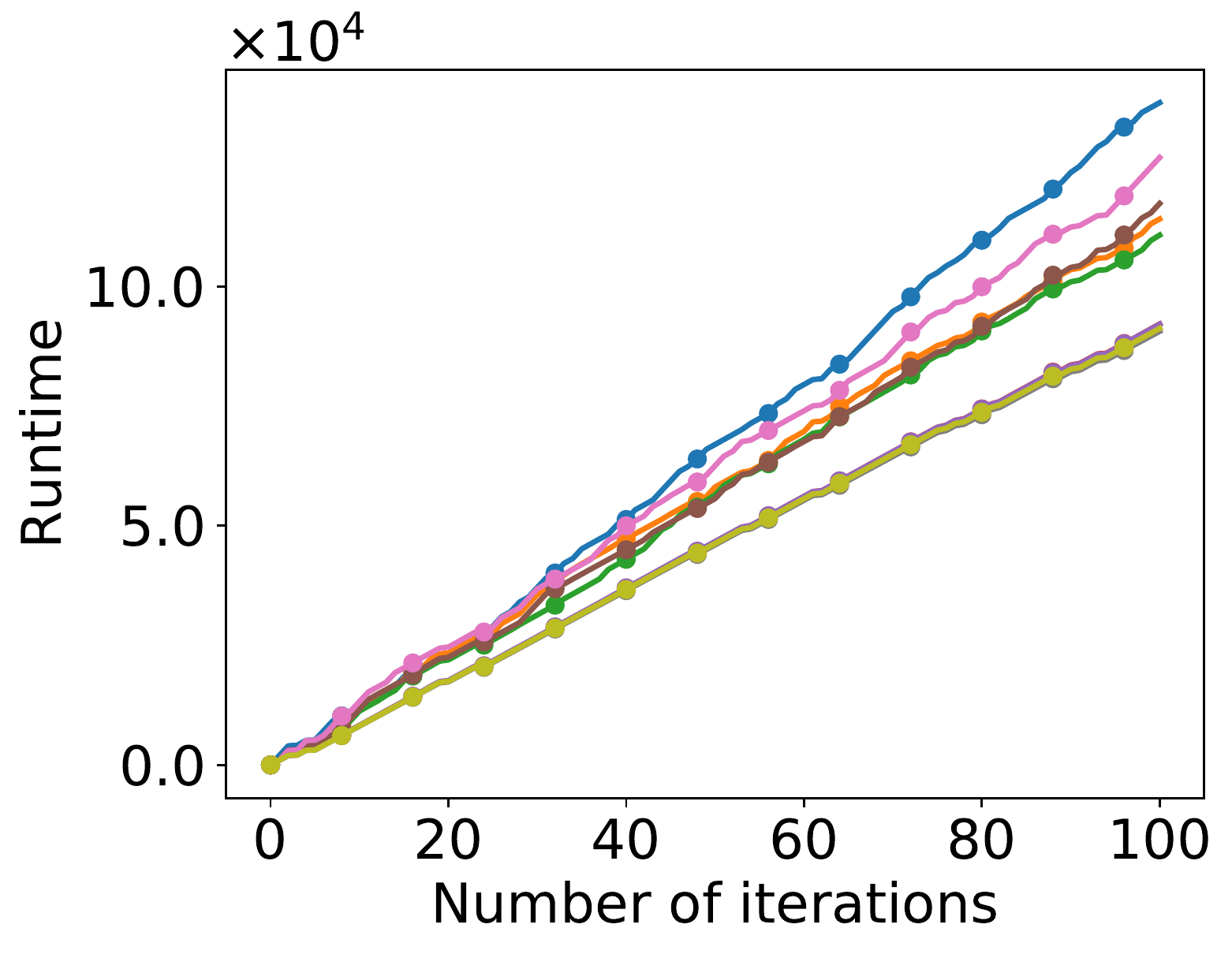}
		\caption{$K = 10$}
	\end{subfigure}
	\begin{subfigure}[b]{0.205\linewidth}
		\includegraphics[width=\linewidth]{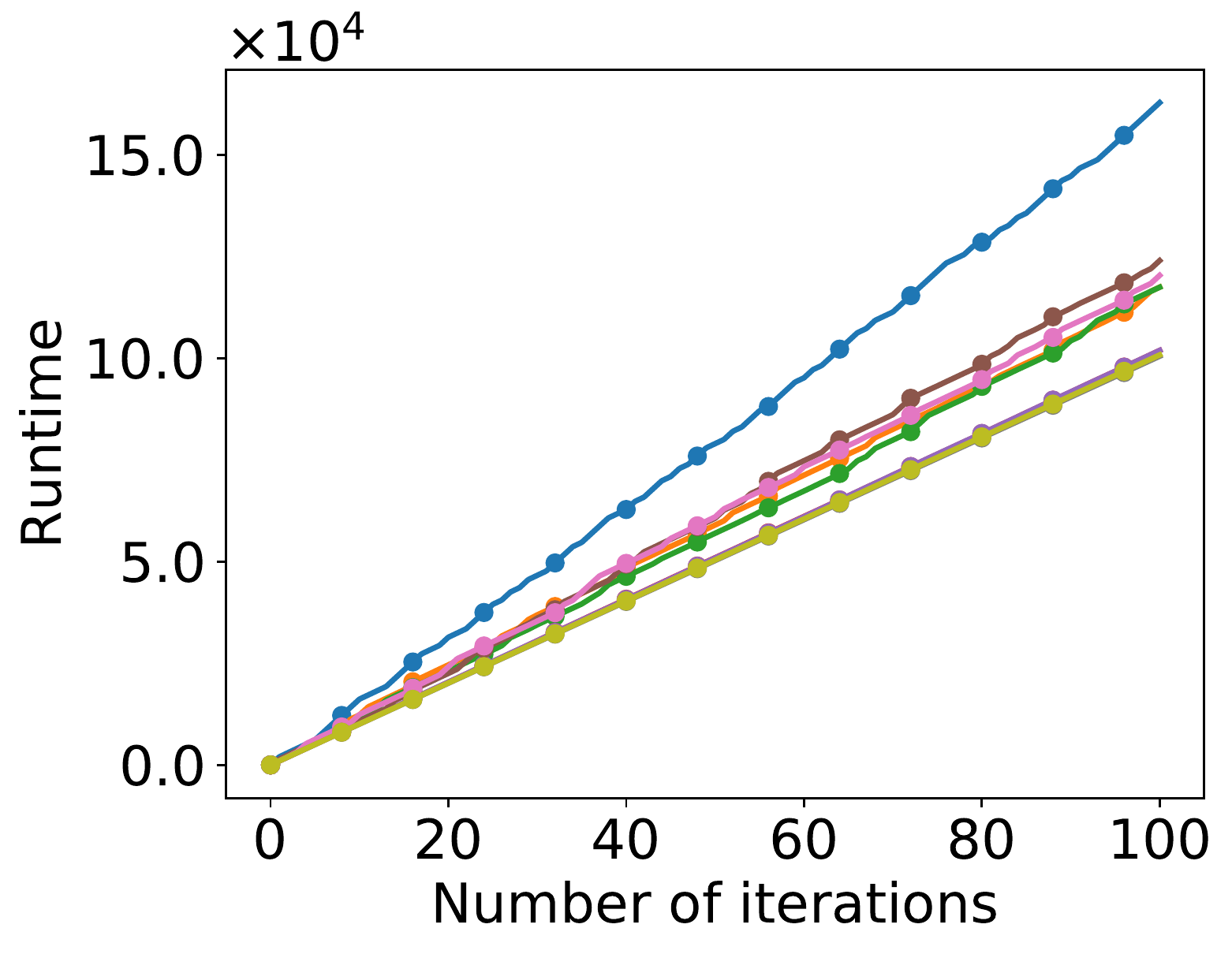}
		\caption{$K = 50$}
	\end{subfigure}
	\vspace{-0.05in}
	\caption{Wall-clock time vs. number of rounds for $N = 1000$ and 4 types of devices in the network.}
	\label{fig:N1000_d4}
	\vspace{-0.2in}
\end{figure*}
\subsection{Settings}
\textbf{ML Model settings.} For the learning task, we choose the textbook handwritten digit recognition task on the MNIST dataset \cite{lecun1998gradient} with 60K training images of $28\times28$ pixels and ten labels from 0 to 9. We consider a 2-layer neural network with a sigmoid activation function in each layer and the hidden layer having a size of 64. The common multi-class cross-entropy loss is applied. For local optimization, we use mini-batch SGD with a batch size of 128. Since our framework is general, the model choices and optimization method are representatives, and the results apply straightforwardly to other settings of ML models and local optimizations.

\textbf{Network settings.} We examine various aspects of the network in use:
\begin{itemize}
	\item Network size ranges from 100 to 1000, resembling small to moderately sized networks. The number of participants in each round is in a typical range $[1:50]$.
	\item Device heterogeneity: Since we have five different devices operating in our EdgeAI testbed (Fig.\ref{fig:edgeai}), we consider two scenarios with, respectively, two and four different types of devices running concurrently. Each device is randomly assigned to one of the available types. We also simulate different stress levels for each device by putting workloads to occupy 25\%, 50\%, and 75\% of total resources, i.e., CPU, memory, and network bandwidth. When a device participates in a round of optimization, a stress level is randomly drawn uniformly.
	\item Communication latency: Since all of our devices in the EdgeAI testbed are placed in the same location, we need to simulate the latency of sending model parameters between devices. We first embed all of them in a 2d plane with coordinates randomly generated in the range $[0,1000]$ and use the distance between two devices to measure the expected latency. An individual transmission's actual latency is drawn from a distribution with the mean value being the expected latency. To determine which distribution to use, we measure latency between a 5G device and a cloud server over the 5G commercial network described in Section \ref{subsec:5gtestbed} and find the best-fitting distributions to experimental latency values. We observe the best fit of a Generalized Extreme Value distribution with parameters of shape $0.7367$ and scale $2.0676$ to communication latencies, as shown in Figure~\ref{fig:PDFlatenciesGEV}. In our experiments, we use the same values of shape and scale but different location values depending on the distances. Note that the network is fully connected. 
\end{itemize}
\subsection{Results}
We measure the time it takes to perform a number of FL rounds when each of the master node selection algorithms is used in our framework. Time is wall-clock and includes both communication and computation times. Here we only measure the time and not the performance because, with the same number of rounds, the performance would be precisely the same for all master node selection algorithms. We also include the time it takes for the optimal master node selections, denoted by \optimalk{} and \optimaln{}, respectively, which essentially try every candidate node and pick the one with the least amount of time at each FL round. Our results are demonstrated in Figures~\ref{fig:N100_d2}, \ref{fig:N1000_d2}, \ref{fig:N100_d4}, and \ref{fig:N1000_d4}. The \textsf{Fixed} curves refer to the central server setting where the fixed server is randomly chosen within the network only for aggregation.

\textbf{Comparison to the (fixed) central server setting:} Our first observation is that the central aggregation server setting requires substantially more time than other master node selection algorithms. For example, it takes up to 4 times more than the \optimaln{}, \stressn{}, and \pown. This difference gets more prominent as the number of participants in an FL round becomes smaller as the fixed server does not consider the locations of participating devices. The group of random master node selections requires marginally less time than the fixed server setting. However, they are still far from optimal for the same reason as for the fixed server.

\textbf{Comparison to the optimal selections:} The second important observation is that master selection algorithms based on least stress with gossip protocol and proof-of-work require mostly the same amount of runtime as the optimal selections for the same number of FL rounds. Particularly, time of \stressk{} and \powk{} are similar to that of \optimalk. \stressn{} and \pown{} are similar to \optimaln{}. This is because both the gossip algorithm and proof-of-work puzzle are swift, i.e., a few seconds for a run \cite{rossi2014distributed,gervais2016security}, compared to the local optimization time. Furthermore, for $K \geq 5$, there is no visible difference in time consumption between the group that selects master among $K$ participants and the group that selects among all $N$ devices in the network.

\textbf{Comparison between different experimental settings:} We also observe an interesting pattern that, as the number of participants $K$ becomes larger, the difference gap between the least and most time-consuming approaches gets smaller. This pattern is intuitive following the law of large numbers since participants are drawn randomly in the network. On another dimension, as the size of network $N$ grows from 100 to 1000, the time each approach takes to perform the same number of FL rounds gets smaller, possibly due to denser network and, thus, the distance between two random devices is lesser. One last thing to note is when the number of device types increases from two (Figures~\ref{fig:N100_d2}, \ref{fig:N1000_d2}) to four - more diverse environments (Figures~\ref{fig:N100_d4}, \ref{fig:N1000_d4}), all the curves exhibit smooth increment of time as the optimization progresses.

\section{Concluding Remarks}
We proposed to select an aggregation server dynamically, termed \emph{flying master}, at each round of optimization in federated learning, compared different metrics to determine the master node, and assessed algorithms to perform the selections. Our experiments on real-world EdgeAI and 5G Testbeds show a significant reduction in runtime when our algorithms are employed, compared to the original FL setting.

\bibliographystyle{IEEEtran}
\bibliography{ref}

\end{document}